\pdfoutput=1
\documentclass[11pt,a4paper]{article}
\usepackage[hyperref]{emnlp2020}

\usepackage{times}
\usepackage{latexsym}
\usepackage{graphicx}
\usepackage{multirow}
\usepackage{gb4e}
\noautomath

\usepackage{microtype}
\aclfinalcopy 

\title{The role of context in neural pitch accent detection in English}

\author{
    {Elizabeth Nielsen~~~ Mark Steedman~~~ Sharon Goldwater}\\
    School of Informatics\\
    University of Edinburgh, UK \\
    {\tt e.k.nielsen@sms.ed.ac.uk} \\
    {\tt\{steedman, sgwater\}@inf.ed.ac.uk} 
}

\date{}

\begin{document}
\maketitle
\begin{abstract}
Prosody is a rich information source in natural language, serving as a marker for phenomena such as contrast. 
In order to make this information available to downstream tasks, we need a way to detect prosodic events in speech.
We propose a new model for pitch accent detection, inspired by the work of \newcite{Stehwien2018}, who presented a CNN-based model for this task. 
Our model makes greater use of context by using full utterances as input and adding an LSTM layer.
We find that these innovations lead to an improvement from 87.5 percent to 88.7 percent accuracy on pitch accent detection on American English speech in the Boston University Radio News Corpus, a state-of-the-art result.
We also find that a simple baseline 
that just predicts a pitch accent on every content word yields 82.2 percent accuracy, and we suggest that this is the appropriate baseline for this task.
Finally, we conduct ablation tests that show pitch is the most important acoustic feature for this task and this corpus.
\end{abstract}

\section{Introduction}

Prosody is a rich information source with the potential to improve performance in many spoken NLP tasks \cite{roesiger2017,Niemann1998}. 
In order to make prosodic information available to downstream tasks, many models have been proposed to predict which words in an utterance carry \textit{pitch accents}---word-level prosodic prominences signaled by a deviation from the speaker's usual pitch, duration, intensity, or some combination of these three features. 
Identifying pitch accents is helpful since they are often used to signal important or unexpected information. 
For example, pitch accents in English typically fall on content words, which are generally more informative. 
When a pitch accent falls on a function word, it indicates that it is unusually informative, as in the sentence, \textit{They ran out of toilet paper even \textbf{before} the quarantine}, where \textit{before} is more informative because it contrasts with what might be a default assumption (e.g., \textit{during}). 

Previous pitch accent prediction models include rule-based models \cite{brenier2005}, traditional machine learning models \cite{wightman1994,levow2005context,gregory2004using}, and neural models \cite{Fernandez2017,Stehwien2017,Stehwien2018}.
\newcite{Stehwien2017} and \newcite{Stehwien2018} (henceforth, SVS18) showed that neural methods can perform comparably to traditional methods using
a relatively small amount of speech context---just a single word on either side of the target word.
However, since pitch accents are deviations from a speaker's average pitch, intensity, and duration, we hypothesize that, as in some non-neural models (e.g. \citealt{levow2005context,rosenberg2009}), a wider input context will allow the model to better determine the speaker's baseline for these features and therefore improve its ability to detect deviations. In addition, we hypothesize that a recurrent model (rather than the CNN used by SVS18) will also improve performance, since it is better adapted to processing long-distance dependencies. 

In this paper, we test these hypotheses by building a new neural pitch accent prediction model that takes in prosodic speech features, text features, or both.
Our main contribution is showing that these context-enhancing innovations in the speech-only model improve performance on a corpus of American English speech, yielding higher accuracy than SVS18 and all previous models on this dataset.
We also find that a baseline of simply labeling all content words with pitch accents is very robust, matching the performance of the text-only model.
We argue that this more robust content-word baseline is the correct baseline for this task.
We find that our speech-only model is able to outperform this baseline by detecting some of the cases where a speaker deviates from the predictions of the content-word baseline, and we provide an analysis of which acoustic features yield the most benefit.  

\section{Models}

\begin{figure}
  \begin{center}
  \vspace{-4.5ex}
    \includegraphics[width=0.8\linewidth]{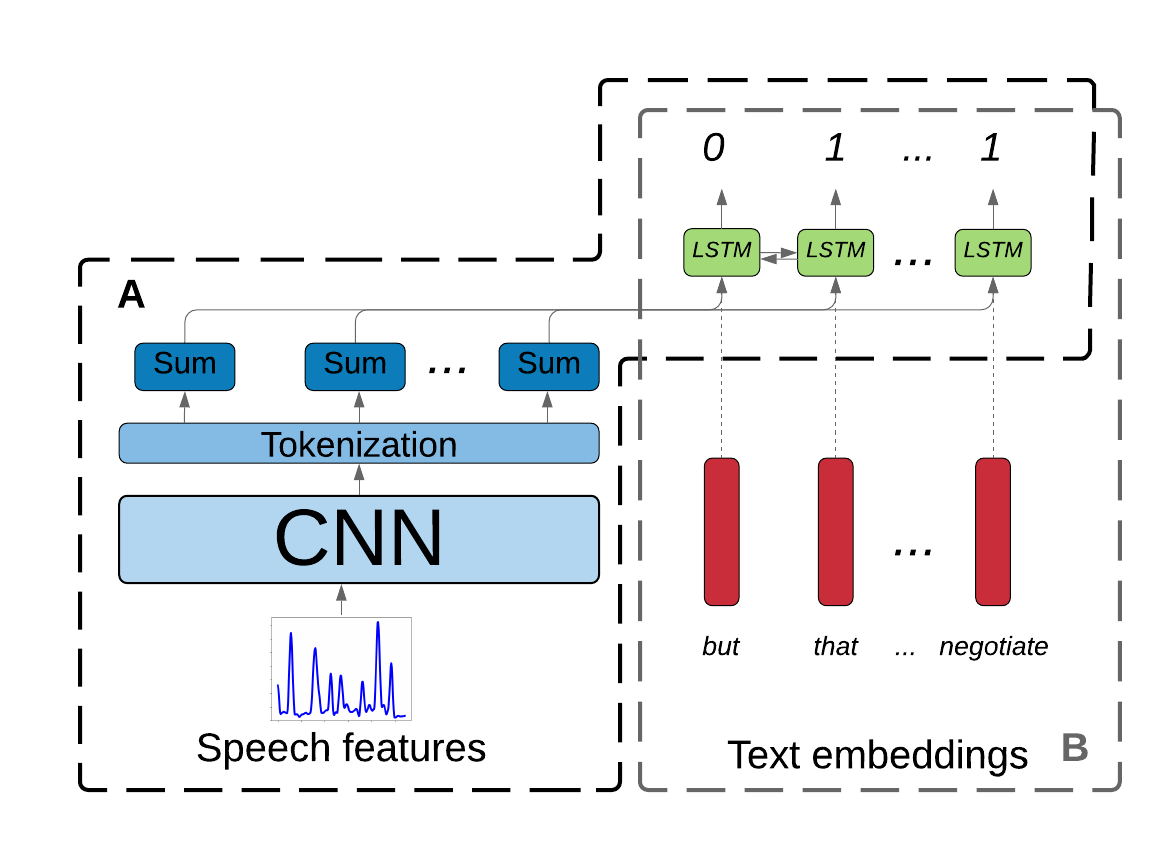}
  \end{center}
     \vspace{-1.5ex}
\caption{ The combined speech+text model. Box A outlines the speech-only model components, while box B outlines the text-only model.} \label{model}

\end{figure}

We build models to predict which words carry a pitch accent, given input of either prosodic speech features, text features, or both. The variants are shown in Figure~\ref{model} and described below. All models are implemented in PyTorch \cite{pytorch}.\footnote{\href{https://github.com/ekayen/prosody\_detection}{https://github.com/ekayen/prosody\_detection}}

We also describe the ways in which we varied the amount of context available to the speech-only model in particular.

\textbf{Speech-only model.} Like SVS18's model, our speech encoder begins with several CNN layers that take a series of frames $f_1, f_2, ..., f_n$ as input, where each frame $f_i$ is a vector of 6 acoustic-prosodic features (see \S \ref{data}). 
These frames are encoded by the CNN, which reduces the overall number of frames by passing a kernel over the input with a stride of size 2, resulting in frames $f'_1,f'_2,...,f'_k$. 
However, rather than predicting the label for a single token at a time, as SVS18 do, our model labels the whole sequence at once. In order to divide the output of the CNN into word tokens, we use the token timestamps provided in the corpus to divide the frames at places corresponding to word boundaries in the input, similar to the approach taken in \newcite{tran2018}.
Each resulting subdivision of the frames $[f'_i,f'_{i+1},...,f'_j]$ contains different numbers of frames, since tokens are of various lengths. 
To obtain token representations of identical size, we sum across all frames for a given token: $t_j = sum(f'_i,f'_{i+1},...,f'_k)$.
Each token embedding $t_1,...,t_m$ is then passed into a bidirectional LSTM, and finally a feed forward layer that outputs a prediction for each token. 
The model's hyperparameters are described in detail in Appendix \ref{hparamappx}.

Our full model takes an entire utterance as input and predicts all labels at once, but we also experiment with using only three or one token(s) as input. 
In these cases, the model only predicts the label for the central input token. 
The three-token scenario is designed to be most similar to SVS18's model.

\textbf{Text-only model.}
The text-only model is a simple bidirectional LSTM. 
An embedding for each token is passed to the BiLSTM and a prediction is made at each timestep. 
We followed SVS18 in using pretrained 300d GloVe word embeddings \cite{glove}, although using pre-trained embeddings did not improve performance much 
over randomly initialized embeddings. 

\textbf{Speech+text model.}
The speech-only and text-only models both include a bidirectional LSTM, so for the combined model, we just concatenate the embedding for each token generated by the CNN encoder with the pretrained text embedding for that token before passing them to the LSTM.

\textbf{Baselines.} In addition to a majority class baseline, we also report results on a content-word baseline, where all content words (non-stopwords as identified by NLTK) are labelled as carrying a pitch accent. We also report a duration-only baseline, where the input features to the speech-only model are all replaced with the value 1---so the model can only tell how many frames each token contains.

\section{Data and experimental setup} \label{data}

We train and test all models using data from the Boston University Radio News Corpus (hereafter BURNC)\footnote{\href{https://catalog.ldc.upenn.edu/LDC96S36}{https://catalog.ldc.upenn.edu/LDC96S36}}, a speech corpus of General American English that is partially annotated with prosodic information. 
The annotated subsection of the corpus that we use includes five speakers, three female, and two male, all of them trained radio journalists reading pre-written news segments. 
This kind of read speech from trained speakers is different from spontaneous speech and so the conclusions we reach here can only confidently be applied to this genre. 
The data we use amount to approximately 2.75 hours of speech, consisting of 1721 utterances. 
These come from a total of 398 news segments. 
There are 28,489 word tokens, 15,544 of which carry pitch accents. 
Though this is a limited amount of data, this corpus is one of very few corpora with available prosodic annotations and enables us to compare with previous studies that use this resource, including SVS18.

For the speech-only model, we follow SVS18 in using the OpenSMILE toolkit \cite{eyben2013} to extract six features, which fall into three broad categories: pitch features (smoothed F0), intensity features (RMS energy, loudness), and voicing features (zero-crossing rate, voicing probability, and harmonics-to-noise ratio). 
These features are extracted from frames of varying sizes (following  \newcite{schuller2013}), and frames are offset by 10ms.
The speech-only model has no access to phone-level or spectral information that might allow it to make predictions based on word identity. 
The transcription of the speech in this corpus includes marked breaths, which we use to segment the corpus into utterances.
Note that there are no explicit correlates of duration in this feature set, though the model has access to the absolute duration of each token via the number of input frames per token. In future, we could follow \newcite{tran2018} by giving an explicit feature for the duration of a given token normalized by the average duration of that token in the corpus. 

For the text-only model, we follow SVS18 in removing contractions (e.g. \textit{we'll} $\longrightarrow$ \textit{we}), though we diverge in leaving hyphenated tokens in place (e.g.\ \textit{eighty-eight} remains \textit{eighty-eight}).

We perform tenfold cross-validation of all experiments and report the average performance. 
For a detailed description of how we divided data into train, development, and test sets for cross-validation, see Appendix \ref{crossval}. 
In order to test for repeatability, we furthermore initialize our model architecture with five distinct random seeds and repeat the tenfold cross-validation procedure for each of these five model initializations. 
Our reported test set results are the average performance of all these five model initializations. 
We report accuracy as our primary metric since this task is a balanced binary classification task. 
We train for 25 epochs and we report the highest development set accuracy of these 25 epochs.
We use this same epoch to report test set accuracy.

\begin{table}
\begin{tabular}{llc} 
     \hline
    \textbf{Context} & \textbf{Architecture} & \textbf{Acc (\%)} \\  
     
     \hline
      
      \multirow{2}{*}{Full utterance}     & CNN+LSTM & 89.1 \\
                                        & CNN only & 87.9\\ 
     \hline
      \multirow{2}{*}{Three tokens}       & CNN+LSTM & 88.6 \\
                                        & CNN only & 87.3 \\
     \hline
     One token                          & CNN only & 85.5 \\
     \hline
    \end{tabular}
    \caption{
    Development set accuracy of speech-only model variants using different input contexts and architectures. Greater input context helps, and including LSTM layers works better than just CNN layers.
    }
    \label{contextresults}
\end{table}

\begin{table}
     \begin{tabular}{lccc} 
     \hline
     & \textbf{Speech} & \textbf{Text} & \textbf{Sp+text}   \\ [0.5ex] 
     \hline
      Our model  & \textbf{88.4} & \textbf{82.2} & \textbf{89.1} \\ 
     SVS18 & 87.1  & 78.5 & 87.5 \\ 
      Content-word & & 82.2 & \\
      Duration-only & 81.2 & & \\
    
     \hline
    \end{tabular}
    \caption{Test set accuracy of our CNN+LSTM model, compared to the CNN-only baseline of SVS18, a baseline where all content words are labelled as accented, and a baseline where the speech model is given only duration information. The majority baseline performance is 54.4 percent.} 
     \label{primaryresults}

\centering
\end{table}

\section{Results and discussion}

Development set results from the speech-only model using different input contexts and architectures are shown in Table~\ref{contextresults}. 
The results confirm our hypotheses that it should help to include more input context (full utterances rather than only three tokens as in SVS18) and to use an LSTM to permit better use of that context. 
Note that our full utterance CNN-only model actually has more parameters than the CNN+LSTM model ($\sim$14m, vs.\ $\sim$12m), so the improvements of the latter are not due to model size. 
The underperformance of the CNN-only model also cannot be attributed to overfitting, since the CNN+LSTM model was more overfit to the training set than the CNN-only model (93.8 percent accuracy vs. 91.7 percent accuracy after 25 epochs).

\begin{figure}\label{textablate}
         \centering
         \scalebox{.6}{
            \includegraphics[width=0.6\textwidth]{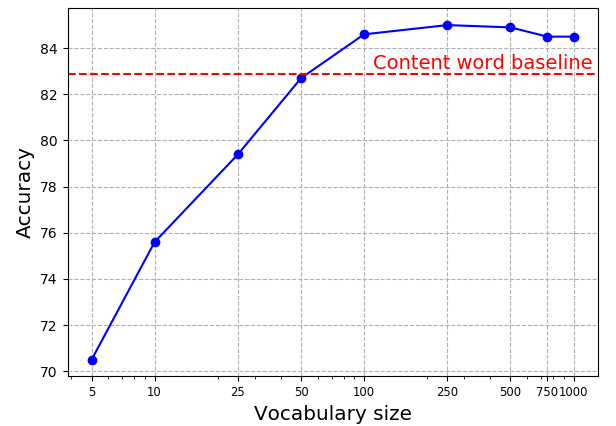}
            }
        \caption{Ablation of vocabulary size in text-only model.}
    \label{fig:vocabshrink}
 \end{figure}

\begin{figure*}
  \centering
        \includegraphics[width=\linewidth]{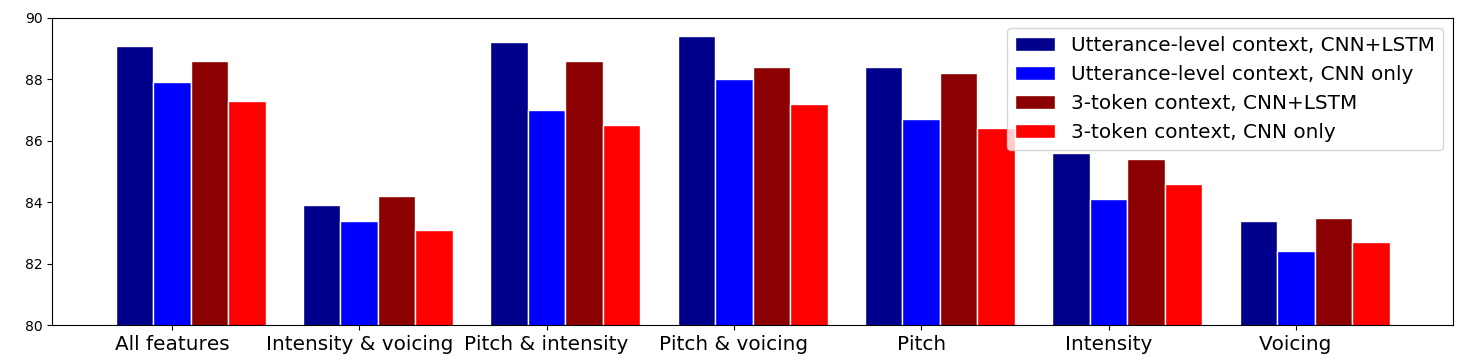}
\caption{Ablation of prosodic features in speech-only model. Labels on the x-axis indicate which features were available to the model.} \label{fig:speechablate}
\end{figure*}

In contrast, development set experiments with the text-only model found little effect of context or architecture (see Appendix \ref{app:devresults}), and indeed even our best text-only model is not much better than the content-word baseline, which in turn outperforms SVS18's text-only model (as shown in Table~\ref{primaryresults} for the test set and Appendix \ref{app:devresults} for the dev set). This suggests that although text-only context might help identify pitch accented words in principle, even powerful neural models are not well able to exploit the right information (or perhaps require discourse level context, which we did not provide). This conclusion is further supported by an additional analysis where we progressively reduced the vocabulary size of the text-only model from 3000 down to 5. As shown in Figure~\ref{fig:vocabshrink}, we found that performance was steady until vocabulary dropped below 100 words (with the rest labelled as `UNK'). This strongly suggests that either word frequency or the strongly correlated content/function word distinction are the main source of information for the text-only model. Of course, absolute word duration is also strongly correlated with frequency and content/function, and we note that the duration-only speech model also achieves a similar accuracy to the content-word baseline (Table~\ref{primaryresults}).

Overall, our best speech-only model outperforms the previous work (SVS18) as well as the text-only model and baselines on the test set (See Table~\ref{primaryresults}), and combining speech plus text yields a small additional improvement.
Our analysis shows in particular that the speech-only model outperforms the text-only model in places where the speaker's realization deviates from the content-word baseline: the speech-only model can correctly detect some pitch accents that fall on function words (as in (\ref{accfunc}) \textit{that}; pitch accents are labeled as \textit{1}) or unaccented content words (as in (\ref{acccont}) \textit{Mary}). 
\begin{exe} 
      \ex
            \begin{xlist}
              
              \ex \label{accfunc} \glll Input: \textit{but} \textit{that} \textit{would}  \textit{require} \textit{the}     \textit{union} \\
                    Speech: 0 1 0 1 0 1 \\
                    Text: 0 0 0 1 0 1 \\ 
            \ex \label{acccont} \glll Input: \textit{she} \textit{agrees} \textit{with} \textit{Mary} \textit{Conroy} \\
                 Speech: 0 1 0 0 1  \\
                 Text: 0 1 0 1 1 \\
            \end{xlist}
\end{exe}

If we only consider these tokens where the speaker's production deviates from the content-word baseline, the speech-only model achieves 66.7 percent accuracy, vs.\ only 38.2 percent for the text-only model.

\begin{table}
\centering
     \begin{tabular}{lcc} 
     \hline
      &\textbf{Our model} & \textbf{Stehwien \& Vu 2017} \\
      & (speech + text) & (speech only)
      \\ [0.5ex] 
    \hline
     f1a & \textbf{89.43} & 85.6 \\
     f2b & \textbf{88.14} & 82.9 \\
     f3a & \textbf{89.65} & 83.5 \\
     m1b & \textbf{85.05} & 81.4 \\
     m2b & 84.42 & \textbf{84.8} \\
     \hline
    \end{tabular}
    \caption{Speaker-independent results of the speech+text model, identified by speaker IDs in BURNC. We compare to the speech-only model of \newcite{Stehwien2017}.} 
     \label{indep}

\centering
\end{table}  

In addition to the evaluations described above, where all utterances are randomly assigned to train, development, and test sets, we do speaker-independent evaluation of the speech+text model.
That is, we hold out a single speaker for testing and use all the other speakers for training and development. 
These results are shown in Table \ref{indep}. 
 We do not have published results of a speech+text model evaluated in this test condition to compare to. 
 However, we can compare to the results of the speech-only model of \newcite{Stehwien2017}. 
We find that our model outperforms theirs on all speakers except the speaker identified as \textit{m2b}. 
The reasons for this underperformance are unclear.

\subsection{Speech feature ablation tests} \label{ablate}

The duration-only baseline shown in Table \ref{primaryresults} shows that the speech model is able to perform quite well given only information about token length, without access to prosodic features, but that these prosodic features are still used in achieving the speech-only model's performance.

In order to determine the relative importance of various prosodic features, we group the prosodic features into those related to pitch (smoothed F0), intensity (RMS energy, loudness), and voicing (harmonics-to-noise ratio, zero-crossing rate, voicing probability), and ablate one or two sets of features at a time. 
We test these models by training them with full utterance context and with more a limited three-token context, as well as with the full CNN+LSTM architecture and the more limited CNN-only architecture. 
The results of these experiments on the development set can be seen in Figure~\ref{fig:speechablate}.

Pitch seems to play the biggest role of these features, with its ablation leading to the lowest performance in all cases.
Voicing appears to be the weakest feature set, actually harming model performance in one case: intensity and voicing features combined underperform intensity features alone.

All three groups of prosodic features seem equally dependent on the inclusion of context, with the removal of the LSTM and restriction to a three-token context leading to proportionally similar drops in performance. This supports our hypothesis that acoustic correlates of prosody cannot be evaluated in isolation: a high pitch or intensity is only meaningfully high compared to some lower pitch or intensity.

\section{Conclusions}

This work demonstrates some important principles for predicting pitch accent from text and speech. 
First, we show that a speech-only model benefits from having utterance-level context. 
Second, we show that both the text and the speech-only model derive at least some of their performance from being able to distinguish function words from content words. 
In fact, our BiLSTM-based text model can hardly outperform a content-word baseline.
Finally, we show that a speech-only model can successfully predict pitch accent in cases where a text-only model cannot, and that combining text and speech provides only a tiny benefit. These results indicate that the speech-only model uses information available in the prosodic features to surpass the content-word baseline, and that knowing the actual words doesn't provide much further useful information.

\section*{Acknowledgments}

We would like to acknowledge the helpful comments of Sameer Bansal, Adam Lopez, and the members of the AGORA research group at the University of Edinburgh. The project SEMANTAX has received funding fom the European Research Council (ERC) under the European Union’s Horizon 2020 research and innovation programme (grant agreement No. 742137).

\bibliography{emnlp2020}
\bibliographystyle{acl_natbib}

\newpage

\appendix

\section{Appendices}
\label{sec:appendix}

\subsection{Development set results} \label{app:devresults}

\begin{table}[ht]
    \centering
    \begin{tabular}{cccc}
    \hline
         & Speech & Text & Sp+text  \\
             \hline
        Our model & 89.1 & 84.5 & 89.8 \\
            \hline
    \end{tabular}
    \caption{Development set results for the full-utterance model.}
\end{table}

\begin{table}[ht]
    \centering
    \begin{tabular}{ccc}
    \hline
          Context & Architecture & Acc (\%)  \\
             \hline
        \multirow{2}{*}{Full utterance} & CNN+LSTM & 84.5  \\
                        & CNN only & 84.4\\
            \hline
         \multirow{2}{*}{Three tokens} & CNN+LSTM & 83.8\\
                    & CNN only & 82.8\\
            \hline
         One token & CNN only & 84.3 \\
            \hline
    \end{tabular}
    \caption{Text model development set results with different context and architectures.}
\end{table}

\subsection{Cross-validation procedure} \label{crossval}

The process we used for tenfold cross-validation was as follows. If we had a corpus with a total of 100 utterances, we would shuffle the utterances, and designate utterances 1-10 as the test set. From the remaining 90 utterances, we randomly designate 10 as development and 80 as training, which gives us our first train/development/test split. Next, we select utterances 11-20 as the test set, and select the development and training sets from the remaining 90 utterances. We repeat this process till we have created 10 distinct train/development/test splits, each with a unique test set. To cross-validate a model, we train and evaluate it on all 10 of these data splits. We use the development portions to optimize hyperparameters, as well as to determine where to stop training for each split.

\subsection{Model hyperparameters} \label{hparamappx}

\begin{table*}[ht]
    \centering
    \begin{tabular}{lcc}
    \hline
       Hyperparameter  & Possible values & Selected values\\
    \hline
      CNN layers   &  2, 3, 4 & 3\\
      LSTM layers   & 2, 3 & 2\\
      Dropout   & 0, 0.2, 0.5, 0.7 & 0.5\\
      Weight decay   & 0, 10e-5, 10e-4 & 10-e5\\
      Filter width   & 9, 11, 13, 15, 17, 19, 21, 23 & 11\\
      Post-tokenization & sum, max & sum \\
      \hline
    \end{tabular}
    \caption{Possible and selected values for each hyperparameter considered in the search. The `post-tokenization' hyperparameter corresponds to the method used to collapse the token representations\,---\,max pooling or summing across all frames.}
    \label{tab:hparam}
\end{table*}

We train the model for a total of 25 epochs of approximately 1400 training examples, using a batch size of 64.
The speech and speech+text models take a total of about 200 seconds to train on average, with each epoch taking around 8 seconds to train on a single Titan X-equivalent GPU. 
The text model takes about 75 seconds to train, with an average of 3 seconds per epoch. 
Evaluation on the entire development set (about 200 instances) takes an average of 2 seconds to run for the speech and speech+text models, and 1 second for the text model.

We perform a hyperparameter search on the combined model, using the resulting hyperparameters for all input configurations (text, speech, speech+text). 
The possible values of each hyperparameter are as shown in Table \ref{tab:hparam}, with each hyperparameter configuration being chosen at random from these values. 
The selected value for the hyperparameter is shown in the right column.
We ran 96 distinct hyperparameter configurations, picking the configuration with the highest accuracy on the development set. 
The average performance on the development set over the search space was 83.5 percent accuracy, with a variance of 0.005 and a standard error of 0.007.

Other hyperparameters are selected manually without searching: we use 128 kernels in the first CNN layer, with 256 kernels in all subsequent CNN layers, and use a stride length of 2 throughout all CNN layers. 
The LSTM layers each have a hidden size of 128. 
We use a learning rate of 0.001 with PyTorch's Adam optimizer \cite{pytorch}.
We set the text-only model to have a vocabulary size of 3000 types, which is approximately 80 percent of the total types present in the corpus.

     \begin{figure}
         \centering
            \includegraphics[width=0.5\textwidth]{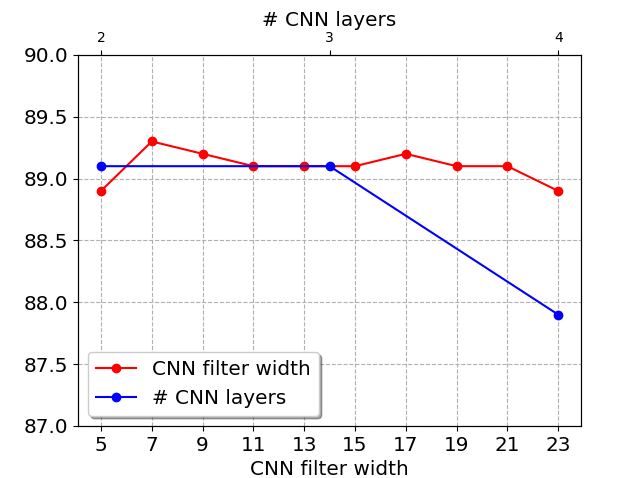}
        \caption{The performance of the speech-only model given different CNN hyperparameters, tested on a development set using tenfold cross-validation. When varying CNN filter width, the CNN layers were kept invariant at 3; when varying the number of CNN layers, the filter width was kept invariant at 11 frames.}
        \label{cnnhparams}
     \end{figure}

Many of our hyperparameter experiments focused on changes to the CNN that should allow it to process a wider swath of the input at once: adjusting filter width, and adjusting the number of CNN layers. Neither change showed significant positive effect, and both were harmful when taken to the extreme. As can be seen in Figure~\ref{cnnhparams}, given a constant depth of 3 CNN layers, the very narrowest kernels underperformed, but widening the kernel did not consistently produce better performance, and eventually degraded performance. Likewise, adding CNN layers---which increases the number of frames of the input data being viewed by the final CNN layer---was actively harmful to performance beyond depths of 3 layers.

\vfill
\ 
\vfill

\end{document}